# Towards Monitoring Parkinson's Disease Following Drug Treatment: CGP Classification of rs-fMRI Data

Amir Dehsarvi

Jennifer Kay South Palomares

Stephen Leslie Smith

# Abstract

**Background and Objective:**

It is commonly accepted that accurate monitoring of neurodegenerative diseases is crucial for effective disease management and delivery of medication and treatment. This research develops automatic clinical monitoring techniques for Parkinson's disease (PD), following treatment, using the novel application of evolutionary algorithms (EAs). Specifically, the research question addressed was: Can accurate monitoring of PD be achieved using EAs on rs-fMRI data for patients prescribed Modafinil (typically prescribed for PD patients to relieve physical fatigue)?

**Methods:**

This research develops novel clinical monitoring tools using data from a controlled experiment where participants were administered Modafinil (typically prescribed for PD patients to relieve fatigue) versus placebo, examining the novel application of EAs to both map and predict the functional connectivity in participants using rs-fMRI data. Specifically, Cartesian Genetic Programming (CGP) was used to classify Dynamic Causal Modeling (DCM) analysis and timeseries data. Results were validated with two other commonly used classification methods (Artificial Neural Networks - ANN, and Support Vector Machines - SVM) and via $k$-fold cross-validation.

**Results:**

Across DCM and timeseries analyses, findings revealed a maximum accuracy of 74.57% for CGP. Furthermore, CGP provided comparable performance accuracy relative to ANN and SVM. Nevertheless, EAs enable us to decode the classifier, in terms of understanding the data inputs that are used, more easily than in ANN and SVM.

**Conclusions:**

These findings underscore the applicability of both DCM analyses for classification and CGP as a novel classification technique for brain imaging data with medical implications for medication monitoring.

Medication monitoring in clinical settings typically does not involve machine learning nor statistical-based methods. Furthermore, classification of fMRI data for research typically involves statistical modelling techniques being often hypothesis driven, whereas EAs use data-driven explanatory modelling methods resulting in numerous benefits. DCM analysis is novel for classification (with research typically using timeseries data) and advantageous as it provides information on the causal links between different brain regions. Hence, the methods examined are less reliant on prior assumptions and provide greater information outputs relative to those commonly used in clinical environments or fMRI research settings.

**Acknowledgements:** The research presented in this paper was part of AD's PhD thesis (University of York). This research involved an analysis of rs-fMRI data acquired from OpenfMRI database (accession number: ds000133; [1]). The participants provided written consent and the study was approved by the ethics committee of University of Chieti (PROT 2008/09 COET on 14/10/2009) and conducted in accordance with the Helsinki Declaration [1], [2].

The authors declare that there are no potential conflicts of interest regarding the research, authorship, and/or publication of this article. Correspondence should be addressed to: Amir Dehsarvi, Department of Electronic Engineering, University of York, Heslington, York YO10 5DD, UK. Email: ad953@york.ac.uk

# 1.    Introduction

By 2050, 22% of the global population will be over 60 years of age, which is double the current proportion (estimated at 11%, 2000). During the early 20th Century, life expectancy in developed nations averaged 50 years of age, whereas in 2014, life expectancy was estimated at an average 80 years of age. This explosion in life expectancy is attributed largely to advances regarding sanitation, medicine, and living conditions [3]–[5]. An ageing population is not without challenges as the biological and cognitive decline associated with ageing is the leading cause of non-communicable diseases, such as cancer, stroke, type-two diabetes, Parkinson's disease (PD) [6], [7] and Alzheimer's disease, amongst others. As such, the World Health Organization's "Draft Twelfth General Programme of Work" (April 13, 2013) lists "Addressing the challenge of non-communicable diseases" [8] as a salient issue. The current research develops a new method of monitoring PD, a neurodegenerative disease (a class of non-communicable diseases), using evolutionary algorithms (EAs).

PD occurs in 0.1-0.2% of the population, in 1-2 % of individuals over 60 years old [9], and it is more common in men [10]. Age is the largest risk factor for PD [7], [11], and increases prevalence by over 400%. Age further increases PD severity to a greater extent than disease duration [6], [11]. PD occurs in all ethnic groups and countries, yet, it is more frequent in Caucasians relative to Asians and Africans [10]. In the United Kingdom (UK), research using data acquired from the National Health Service (NHS) General Practice Research Database[1] (2009) estimated PD prevalence at 0.3% (0.3% for men and 0.2% for women), which corresponds to 126,893 PD patients[2]. Given a globally ageing population, UK prevalence is estimated to raise 28% by 2020.

Modafinil (Provigil) is a multipurpose drug used in the treatment of sleep disorders, such as narcolepsy [12], [13], and it improves cognitive function in psychiatric disorders, for instance schizophrenia [14]–[16]. Modafinil is a fatigue-reducing medication that boosts attention and memory. Approximately half of PD patients report fatigue-related symptoms [17], for which Modafinil is a common medication that clinicians prescribe [17]–[20]. Research has linked

---

[1] World's biggest database of anonymised longitudinal medical records, comprising approximately 3.4 million people's records
[2] Approximate number of cases per country: 108,000 in England, 10,000 in Scotland, 5,900 in Wales and 3,000 in Northern Ireland.

Modafinil to enhanced attention and memory [21]–[23] and it is well tolerated by patients and has no effect on PD movement symptoms [24]. More recently, Modafinil has been used as a "smart drug", with approximately 25% of students at Oxford, Newcastle, and Leeds having tried this drug [25]. In a 2013 interview with the Telegraph, Prof Sahakian of clinical neuropsychology (University of Cambridge) mentioned that the number of students taking Modafinil was increasing alarmingly [26]. There is much controversy regarding the use of psychostimulants or "smart drugs" to heighten cognitive abilities [27]. Research suggests that Modafinil results in less side effects and lower addition risks, relative to stimulants such as methylphenidate [28] and amphetamine [29], with benefits similar to those provided by these drugs. Yet, Modafinil has a large impact on the dopaminergic system, indicating possibly stronger addiction risks than previously estimated [30].

Even though Modafinil is regularly prescribed to PD patients, evidence supporting its effectiveness as a fatigue-reducing drug is mixed. For instance, Lou et al. conducted a longitudinal study testing participants multiple times over two months and revealed that Modafinil can diminish physical fatigue in PD patients, as noted by the Epworth Sleepiness Scale and by a finger-tapping task following two months of treatment [19]. Other research has revealed that Modafinil relieves fatigue, as measured by the Global Clinical Impression Scale for Fatigue, and reduces excessive daytime somnolence [20]. Nevertheless, this same research revealed no relief from fatigue as measured by the Fatigue Severity Scale. A crucial point is that, regardless of the effectiveness of Modafinil in reducing fatigue, it is commonly prescribed to PD patients.

## 1.1.    Learning Algorithms

Regular monitoring every six months of motor and non-motor symptoms via outpatient clinic appointments is necessary to ensure patients are administered appropriate levels of medication, given their changing symptomology (and medication side effects). Unfortunately, it is common for drugs administered to treat one symptom to also aggravate another symptom. For instance, D1 dopamine agonists reduce bradykinesia but worsen hallucinations and levodopa reduces tremor but worsens dyskinesia. Therefore, clinicians need to make a trade-off between the benefits (reducing PD symptoms) and costs (side effects) when prescribing medication. Hence, this research develops a technique for accurate monitoring of PD using EAs on rs-fMRI data for participants prescribed Modafinil.

Specialised equipment, including learning algorithms, are becoming increasingly relevant for monitoring early stages of cognitive and motor decline, providing enhanced reliability and efficiency relative to medics. For instance, a medic typically measures motor decline via a finger-tapping task (patients tap their thumb and forefinger multiple times as widely and as quickly as they can). Yet, this task has poor inter- and intra-rater reliability given that medics can find it difficult to assess performance based only on the observation of patient movement [31]. A novel technique involves applying computational techniques (evolutionary algorithms, EAs) to evaluate patient movements. Research reveals accuracies of ~95% when EAs are used to analyse data from PD patients (versus age-matched controls) completing the finger tapping task [32]. Therefore, the application of novel computational methods to data from commonly used motor assessments is a crucial and relatively new technique in monitoring PD.

The classification literature has examined PD patients using learning algorithms. EAs are a crucial and novel technique relevant for disease classification, including PD patients. EAs are optimising algorithms derived from Darwinian evolutionary theory. Cartesian Genetic Programming (CGP) is a subtype of EAs that by default evolves directed acyclic computational structures of nodes. Recurrent CGP (RCGP) is an extension of CGP that includes cyclic or feedback connections. The application of CGP and RCGP to neuroimaging data (including rs-fMRI, as per this research) has not been explored and potentially present novel and relevant tool for disease monitoring, which is the focus of the current research.

EAs combined with an expressive dynamical representation enable researchers to examine a large variety of classifiers, hence, they can be a crucial and valuable tool. These classifiers extend the current knowledge in the field given that they are efficient in detecting trends without the use of expert knowledge, leading to findings that may not have been identified by experts. For example, evolved classifiers and their distributions have led to the following scientific contributions: diagnostic accuracy is influenced by dominance, specific acceleration movements are abundant in PD patients, and amplitude and frequency blends with diagnostic power. Nevertheless, trends identified by these algorithms can seem incomprehensible to experts as the solutions obtained are not grounded in expert knowledge. Therefore, whilst these classifiers are a relevant in guiding and/or supporting a medical diagnosis or disease management, any monitoring solutions require a clinician/expert to outline possible biological foundations of these solutions.

A crucial overarching research question is: Can methods for differentiating clinical groups using EAs on rs-fMRI data be developed, based on a controlled clinical experiment and focusing on a medication (Modafinil) typically administered to PD patients? This experiment examines rs-fMRI data in which young adults were administered a single dose of 100 mg of Modafinil (versus not). This experiment used healthy participants rather than PD patients in order to initially develop accurate monitoring of individuals administered a common PD medication, with implications for PD patient monitoring. The activity of Resting State Networks (RSN) and the intrinsic connectivity are explored, and these data are subjected to a classifier to examine the physiological impact of Modafinil.

## 1.2.    rs-fMRI Data

There is a broad literature on rs-fMRI [33]–[35], including research examining the classification of participant groups or brain states using functional connectivity. For example, Esposito et al. explored resting state functional connectivity differences in participants administered Modafinil [2]. rs-fMRI enable researchers to examine specific questions that are not suited to task-related fMRI paradigms, focusing on changes in functional connectivity [36] and facilitating the concurrent exploration of multiple cortical circuits. In addition, confounding variables (e.g., between participant variance in task performance) are reduced given that participants are at rest [37].

Rest activation is linked to minimum ten [38]–[41] functional RSN [38], such as the Default Mode Network (DMN) [42], the Salience Network, the Fronto Parietal Control (FPC) network (lateralised in both hemispheres), the primary Sensory Motor Network, the Exstrastriate Visual System, and the Dorsal Attention Network [38]. DMN is typically examined when conducting rs-fMRI research and, therefore, it is explored in the current study. Abnormal patterns of activation can be investigated using functional connectivity, nevertheless, causal influences (effective connectivity, i.e., the causal impact of one neuronal system on another) cannot be determined [43]. Dynamic Causal Modeling (DCM) [43]–[46], in contrast, is derived from a neuronal model of coupled neuronal states and enables investigation of the effective connectivity underlying functional connectivity via nonlinear designs to detect a generative model of measured neural activity [47]. The current research applies EAs, specifically CGP and RCGP, for the classification of rs-fMRI in PD using DCM and timeseries analyses. Findings are validated via two typically used classification techniques (Artificial Neural Networks, ANN, and Support Vector Machines, SVM) including via *k*-fold cross-validation (CV).

This research applies an exploratory, data-driven approach, being the first to examine Modafinil classification to differentiate participant groups. Developing a method to differentiate participants is clinically relevant with implications for patient diagnosis (possible classification of patients relative to healthy controls) and drug treatment monitoring. This research addresses two further novel research questions: (1) Can Cartesian Genetic Programming (CGP) be applied to the monitoring of participants administered Modafinil (versus a control group) using rs-fMRI data? And (2) are timeseries analyses and DCM analyses relevant for classification? CGP and RCGP have not been applied to the classification of brain imaging data. Research, moreover, has not examined the applicability of DCM data for classification with limited research even applying DCM to clinical data [48]–[53]. By addressing these questions, the research develops automatic procedures for locating brain imaging preclinical biomarkers, these being crucial for PD monitoring.

### 1.3.    Class-Imbalanced Samples

This research involved an analysis of data acquired from OpenfMRI, in which there are unequal group sample sizes. Recruiting equal sample sizes of patients and healthy controls is a common challenge in medical research. When there are unequal groups of controls versus patients the data is described as being class-imbalanced. Imbalanced data reduces classification accuracy [54]–[56], given that classification methods typically assume balanced class distributions [57]. Four implications of class-imbalanced data are as follows: (1) the training set sample size is reduced since classifiers typically consider that imbalanced data has limited sample sizes; (2) learning is biased towards classes with bigger sample sizes; (3) classification rules predicting the majority class tend to be specialised, hence, with low coverage, and so general rules (predicting the minority class) are prioritised; and (4) the classifier may incorrectly label small groups of minority class data as noise and reject these. Further, classification of the minority class can be worsened by the presence of genuine noise data, given the small sample size available, as there is less accurate data available with which to train the classifier [58].

Solutions commonly involve changing/generating data to obtain more balanced class-distributions, resulting in heightened classification accuracy [59]–[61]. This research applies ADASYN [62] to generate synthetic data leading to balanced class-distributions for the training set. Nevertheless, a limitation of generating synthetic balanced data is that it is applied only to the training sets. The classifier trained for CV is a different classifier with a different accuracy level than that used in the classification part. Therefore, classification accuracy for the validation and test sets is typically reduced. Indeed, the development of accurate classifiers that can be applied to heavily class-imbalanced data with the objective of detecting a minority class (e.g., PD patients) are required.

## 1.4.    Research Overview and Aims

The current research adopts a novel and exploratory approach to develop novel clinical monitoring tools using data from a controlled experiment in which participants were administered the drug Modafinil, typically prescribed for PD patients to relieve physical fatigue. Specifically, this experiment explored the question: Can accurate monitoring of PD be achieved using EAs on rs-fMRI data for patients prescribed Modafinil? This research involved an analysis of rs-fMRI data taken from OpenfMRI database (accession number: ds000133; [1]), in which healthy young adults were administered a single dose of 100 mg of Modafinil (versus not). In the current research, the activity of the RSN and the functional connectivity were examined, and this data was subjected to classification to explore the physiological impact of Modafinil.

This research applied EAs, specifically CGP and Recurrent CGP (RCGP), for the classification of rs-fMRI using DCM and timeseries analyses. Supervised classification was conducted on the timeseries and DCM analyses from the rs-fMRI data. CGP and RCGP have not hitherto been applied to the classification of brain imaging data. Moreover, this research explores a further novel question: is DCM analysis relevant for classification? Research has not examined the usefulness of DCM analyses for classification.

A common limitation in medical research involves recruiting equal sample sizes of patients and healthy controls, often leading to class-imbalanced data (e.g., unequal groups of controls versus patients). The OpenfMRI dataset used in this experiment is heavily class-imbalanced, with many more control participants relative to participants administered Modafinil. Hence, this research explored the applicability of classification methods to class-imbalanced data, with implications for the transferability of medical research based on limited and imbalanced sample sizes.

## 2.  Methods

### 2.1.  Participants

Twenty-six male participants were tested (age range: 25–35 years). Participants were right-handed (tested using the Edinburgh Handedness inventory) [63], with similar educational level, and no history of psychiatric, neurological or medical (hypertension, cardiac disorders, and epilepsy) conditions as identified by the Millon test and by clinical examination. The participants provided written consent and the study was approved by the ethics committee of University of Chieti (PROT 2008/09 COET on 14/10/2009) and conducted in accordance with the Helsinki Declaration [1], [2].

### 2.2.  Procedure

Participants were told to consume their normal amount of nicotine and caffeine and to refrain from consuming alcohol 12 hours prior to the study. Participants were administered Modafinil (100 mg) or placebo. The study was double blind and both the Modafinil and placebo pills looked identical. Following consumption of the drug, participants were given an fMRI scan.

Data from all participants were processed and separated into their corresponding treatment groups (Modafinil and placebo). Data from one control participant was excluded, as this data was too poor quality to be analysed. Pre-session data was also categorised as placebo/control as this research only examines the effect of Modafinil on brain functionality. The Modafinil group contains 39 participants: 13 participants tested in one session with three runs. The control group contains 111 participants: 12 participants tested in two sessions with three runs plus 13 participants tested in one session with three runs.

## 2.3.    rs-fMRI Acquisition

rs-fMRI BOLD data was separated in three runs (duration: 4 min each) and then high resolution T1 anatomical images were acquired. Participants were instructed to focus on the centre of a grey screen that was projected on an LCD screen, viewed via a mirror located over the participant's head. The participant's head was placed in an eight-channel coil with foam padding to reduce involuntary head movements. BOLD functional imaging was performed with a Philips Achieva 3T Scanner (Philips Medical Systems, Best, The Netherlands), using T2*-weighted echo planar imaging (EPI) free induction decay (FID) sequences and applying the following parameters: Echo Time (TE) 35 ms, matrix size $66 \times 66$, Field of View (FoV) 256 mm, in-plane voxel size 464 mm, flip angle 75°, slice thickness 4 mm and no gaps. 140 functional volumes consisting of 30 transaxial slices were acquired per run with a volume Repetition Time (TR) of 1671 ms. High resolution structural images were acquired at the end of the three rs-fMRI runs through a 3D MPRAGE sequence employing the following parameters: sagittal, matrix $256 \times 256$, FoV 256 mm, slice thickness 1 mm, no gaps, in-plane voxel size $1\ mm\ \times\ 1\ mm$, flip angle 12°, TR = 9.7 ms and TE = 4 ms [1], [2].

## 2.4.    Imaging Data Analysis

### 2.4.1.  Preprocessing

The imaging data analyses were conducted with CONN (version 17.c) [64] and SPM12 (version 6906 - Wellcome Department of Imaging Neuroscience, London, UK) [65] software packages based on MATLAB. Preprocessing included 4D NIFTI import, 4D to 3D NIFTI conversion, and decrease of spatial distortion via the Field Map toolbox in SPM12 [65]. Anatomical data was segmented and both anatomical and functional data were normalised. All the functional images were motion corrected and coregistered to participants' own high-resolution anatomical image. The participants' anatomical images were normalised to the standard T1 template in the MNI space, as provided by SPM12. Subsequently, the normalisation parameters of each participant were applied to the functional images to normalise all the functional images into the MNI space. The EPI data was unwarped (using field-map images) to compensate for the magnetic field inhomogeneities, realigned to correct for motion, and slice-time corrected to the middle slice. The normalisation parameters from the T1 stream were then applied to warp the functional images into MNI space. All the functional images were spatially smoothed using a Gaussian kernel with 8 mm FWHM [66] to account for inter-participant variability while maintaining a relatively high spatial resolution. Linear and quadratic detrending of the fMRI signal was applied, which involved covarying out WM and CSF signal. WM and CSF signals were predicted for each volume from the mean value of WM and CSF masks, derived by thresholding SPM's tissue probability maps at 0.5. The data was bandpass filtered (0.008–0.1 Hz).

### 2.4.2. Processing

#### 2.4.2.1. Timeseries

Functional connectivity in DMN is well studied. Hence, this research took as regions of interest (nodes) the most commonly reported four major parts of DMN, as shown in Figure 1: medial prefrontal cortex (mPFC, centred at 3, 54, −2), posterior cingulate cortex (PCC, centred at 0, −52, 26), left inferior parietal cortex (LIPC, centred at − 50, − 63, 32), and right inferior parietal cortex (RIPC, centred at 48, − 69, 35). For each participant, the volumes of interest were defined as spheres centred at the coordinates outlined above with an 8 mm [66] radius and with a mask threshold of 0.5. The first eigenvectors were extracted after removing the effect of head motion and low frequency drift. This vector is stored for each region as timeseries.

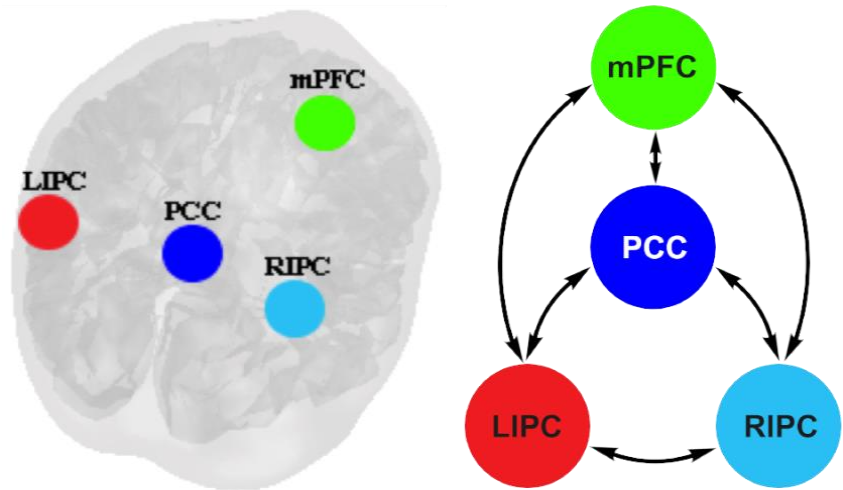

Figure 1 - The four DMN regions of interest used in this research

## 2.4.2.2.        Dynamic Causal Modeling (DCM)

The spectral DCM analyses were conducted using the DCM12 implemented in SPM12. The regions of interest of DCM analyses were defined according to the peak of the DMN independent component maps, as presented in Figure 1. The fundamental aim of this DCM analysis was to examine the endogenous/intrinsic effective connectivity and to investigate the causal interactions across these regions. The modelled low frequency fluctuations were set as key inputs to all four nodes, and different models were defined by considering a full connection for all nodes. Expected posterior model probabilities and exceedance probabilities were computed. The intrinsic connectivity parameters (16 values that were stored in DCM.Ep.A matrix, all parameters of intrinsic/effective connectivity [43]) from each participant were subjected to classification using CGP.

## 2.5.    Cartesian Genetic Programming

CGP [67] is a strand of GP [68], [69], that encodes computational structures as generic cyclic/acyclic graphs. For this research, a new cross platform open source CGP library [70] was used since it is able to evolve symbolic expressions, Boolean logic circuits, and ANN, and it can be extended to diverse research areas. The CGP library enables the control of evolutionary parameters and the application of custom evolutionary stages.

### 2.5.1. Classification

To have equal class representation, data from each class was divided randomly into subsets of 70% (training), 15% (validation), and 15% (test). The geometry of the programs in the population (chromosomes) has fifty nodes with a function set of four mathematical operations (+, -, ×, ÷), multiple inputs (according to the datasets), and one output (class 1 for Modafinil participant group, class 0 for the control participant group). At each generation, the fittest chromosome is selected and the next generation is formed with its mutated versions (mutation rate = 0.1). Evolution stops when 15000 iterations are reached. To obtain statistical significance, the classification was done in 10 runs for each combination of inputs and the accuracy was averaged over the runs. The results (the winning chromosome, the networks, and the accuracy values) were stored for each run individually.

#### 2.5.1.1. Classification of Timeseries

rs-fMRI is a widespread tool for exploring the functionality of the brain, using volume timeseries data. These scans contain abundant data; hence, obtaining relevant and useful data from raw scans (i.e., high dimensional datasets) can be difficult. Machine learning algorithms provide various tools that create datasets with less dimensions and more useful data, although challenges persist regarding how to select relevant data and how to maintain the interpretability of this data. This can result in losing important properties of the raw data, although dealing with such a large number of features can be computationally expensive and very time consuming.

In the current experiment, the RCGP algorithm is used to classify the features that appeared across time in participant scans. The number of timeseries values is 145, i.e., for each of the four regions in the DMN, there is a vector of 145 values. Analysing/classifying the timeseries values was conducted in three different ways in terms of inputs to the classifier:

1. The timeseries values for each region separately were used as inputs to the classifier, to classify with 145 features per region (relating to DMN regions: LIPC, RIPC, PCC, and mPFC) and per participant.

2. The timeseries values together in four columns (one per region) were used as inputs to the classifier to classify the data with four columns (corresponding to the four DMN regions) of 145 features per participant.

3. The timeseries values inserted together in one column were used as inputs to the classifier to classify the data with 580 features in one vector for each participant. The order of inputting the timeseries values for each DMN region to form the final vector was consistent between participants.

Classification was completed in 10 runs per combination of inputs and the mean across the runs was calculated. The same inputs were used to classify the data using ANN and SVM in MATLAB for comparison/validation. RCGP was used with 10% probability for the recurrent connections. A complete pipeline of the preprocessing and processing of the data is presented in Figure 2.

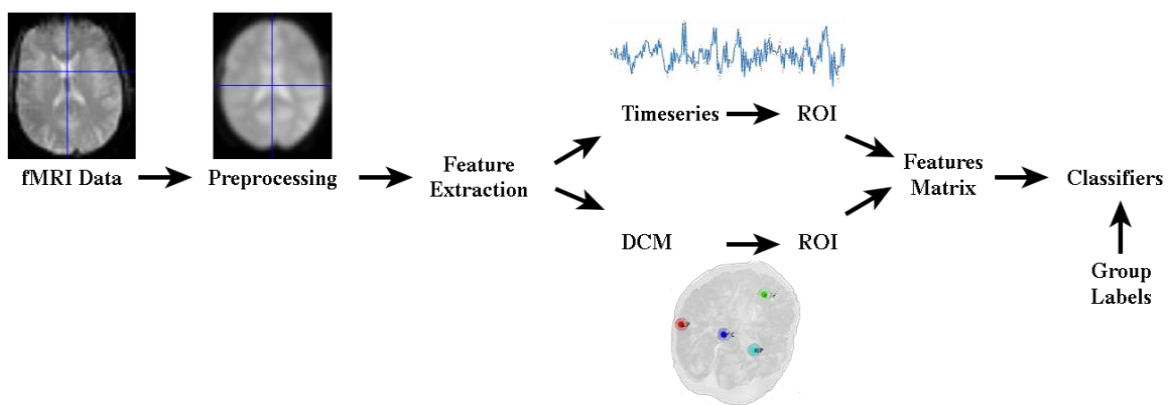

Figure 2 - Data analysis pipeline used in this research

## 2.5.1.2.    Classification of DCM

The classification was run in the CGP Library with 16 inputs (all the DCM values sorted by region and presented as only one vector for each participant). To facilitate comparison, the classification with data from the same participants was run using ANN and SVM, both run in MATLAB.

## 2.6.    Adaptive Synthetic Sampling

As previously mentioned, the Modafinil group contained 39 participants and the control group contained 111 participants, resulting in highly imbalanced data. Hence, ADASYN was used to make the data balanced. After the process, the minor group for each combination (in the training set) had a higher number of participants, which made the data balanced for cross-validation (CV). The validation and test sets were kept the same.

## 2.7.    *k*-Fold Cross-Validation

*10*-fold CV was used to evaluate the classification accuracy using an unbiased estimate of the generalisation accuracy [71]. Advantages of CV include generating of independent test sets with improved reliability. With *10*-fold CV, typically one (of *10*) subset is the test set and remaining nine subsets are training sets. These sets are then rotated so that each set is used to test the data once. One repetition of the *10*-fold CV does not produce sufficient classification accuracies for comparison, therefore, *10*-fold CV is repeated 10 independent times and the mean accuracy over all *10* trials is calculated.

Since the main classification methodology in this research involved dividing the data into three different subsets (training, validation, and test), the data was divided into *10* subsets and, each time, one of the *10* subsets were used as the test set, another one for validation, and the remaining eight as a training set. The data was divided using stratified random sampling enabling the sample proportion in each data subset to be the same as that in the original data (i.e., equal class distribution in the subsets as per the original data). Hence, the data was split for each class initially and then the classes were mixed to form the completed set. This was done for each combination of inputs to both CGP (for DCM values) and RCGP (for timeseries values). In this study, the data for each combination of inputs was divided into three parts of 80% (training), 10% (validation), and 10% (test).

## 3.    Results

This study examined the classification of 39 participants administered Modafinil versus 111 control participants. The analysis (classification) focused on organising features to be used as inputs to the classifier in CGP and also in RCGP, implemented using the CGP Library. To validate the results, the analysis/classification was additionally completed using ANN and SVM, both in MATLAB.

## 3.1. Classification of Timeseries

Initially, the timeseries values for each region were used as inputs to the classifier individually. Therefore, the data was classified with 145 features from each region per participant. The same procedure was completed separately for each DMN region (PCC, mPFC, RIPC, and LIPC). Then, the timeseries values were used as inputs to the classifier to classify the data in four columns (relating to the four DMN regions) of 145 features per participant. Finally, the timeseries values together in one column were used as inputs to the classifier to classify the data with 580 features in one vector for each participant. The results after 10 runs for each combination were averaged and are summarised in Table 1. The classification was also done using ANN and SVM. For SVM, only the training and test sets were considered for classification.

Table 1 - Classification results for the timeseries values

| | | Training % (*SD*) | Validation % (*SD*) | Test % (*SD*) |
|---|---|---|---|---|
| Classification results for each DMN region | | | | |
| PCC | RCGP | 73.75 (2.98) | 75.52 (1.59) | 74.28 (1.22) |
| | ANN | 83.74 (7.78) | 73.48 (12.88) | 72.18 (8.01) |
| | SVM | 100 (0) | NA | 73.81 (0.50) |
| mPFC | RCGP | 71.88 (5.00) | 74.80 (1.57) | 73.52 (1.38) |
| | ANN | 78.07 (9.02) | 71.75 (9.90) | 64.78 (9.93) |
| | SVM | 100 (0) | NA | 74.00 (0.46) |
| RIPC | RCGP | 74.40 (2.96) | 76.51 (2.57) | 71.26 (5.94) |
| | ANN | 81.92 (7.63) | 72.18 (7.46) | 70.00 (8.06) |
| | SVM | 100 (0) | NA | 73.71 (0.49) |
| LIPC | RCGP | 72.79 (3.52) | 74.79 (1.55) | 72.53 (3.22) |
| | ANN | 85.88 (10.20) | 70.88 (11.96) | 70.45 (8.65) |
| | SVM | 100 (0) | NA | 73.90 (0.49) |
| Classification results for all the DMN regions (4 inputs) | | | | |
| | RCGP | 73.63 (2.11) | 75.92 (2.16) | 73.99 (2.94) |
| | ANN | 73.83 (0.17) | 74.49 (0.67) | 74.24 (0.49) |
| | SVM | 74.25 (0.05) | NA | 73.95 (0.05) |
| Classification results for all the DMN regions (1 input) | | | | |
| | RCGP | 74.68 (1.60) | 76.55 (2.82) | 74.57 (1.82) |
| | ANN | 73.97 (0.13) | 74.11 (0.37) | 74.01 (0.45) |
| | SVM | 74.00 (0.00) | NA | 74.00 (0.00) |

Findings revealed that the Modafinil group were successfully classified from the control group with a maximum accuracy of 74.57% using RCGP (minimum accuracy: 71.26%). The results from the other two classification techniques (ANN and SVM) validated this finding as they were comparable: 64.78-74.24% for ANN and SVM in all the different combinations of inputs.

Unlike for the DCM analyses, mixed ANOVAs were not conducted to evaluate the correspondence between participant group and timeseries features, given that there were 580 features per participant, which would not be interpretable.

## 3.2. Classification of Dynamic Causal Modeling (DCM)

Classification using the CGP Library was executed with 16 inputs (all the DCM values sorted by region and presented as one vector for each participant) and 1 output (class 1 for the Modafinil group and class 0 for the control group). The results were then averaged over 10 runs and are presented in Table 2. The classification was done using ANN and SVM. For SVM, only the training and test sets were considered for classification.

Table 2 - Classification results for DCM values

|  | Training % (*SD*) | Validation % (*SD*) | Test % (*SD*) |
| --- | --- | --- | --- |
| CGP | 75.63 (7.13) | 80.35 (6.18) | 73.89 (7.70) |
| ANN | 79.99 (5.57) | 73.93 (7.94) | 67.39 (6.57) |
| SVM | 73.81 (0.50) | NA | 73.81 (0.50) |

Findings revealed that the Modafinil group were successfully classified from the control group with 73.89% accuracy using CGP. The results from the other two classification methods (ANN and SVM) validated this finding as they were comparable: 67.39% for ANN and 73.81% for SVM.

Two examples of the CGP classification network trees/graphs can be seen in Figure 3 and Figure 4. This reflects one of the fundamental gains of CGP in terms of providing a *white box solution*, providing more detailed information on (a) the inputs used, and (b) the final solution generated in classification; this is not easily feasible (if at all) with ANN and SVM classification methods. These networks and their respective mathematical expressions are very complex and often difficult to interpret. Nevertheless, these networks can provide highly useful information. For example, in Figure 3, only half of the inputs have been used to arrive at the final classification. Similarly, Figure 4 reveals that only 11 of the inputs are used and the network evolved, in this case, is arguably somewhat simpler than that depicted in Figure 3. Future work can explore constraining the geometry of the classifier (at the cost of classification accuracy) in order to simplify the networks and enhance interpretability.

Figure 3 - CGP classification tree for the classification of Modafinil vs. control; example 1

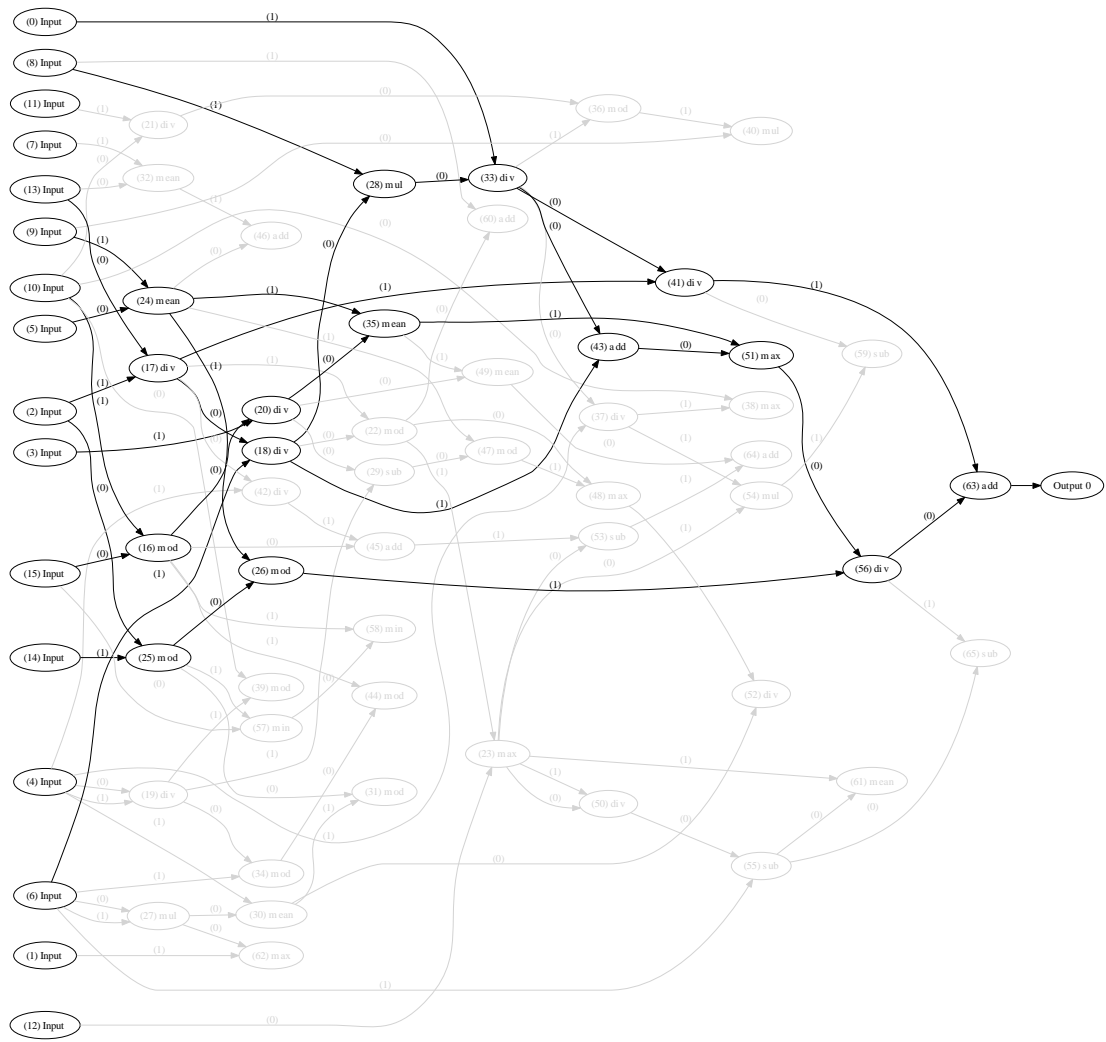

Figure 4 - CGP classification tree for the classification of Modafinil vs. control; example 2

To evaluate the correspondence between participant group and DCM features, a mixed $2 \times 16$ ANOVA between the participant group (Modafinil and control) and DCM features (16 inputs per participant) was conducted. A Greenhouse-Geisser correction was used as the model violated sphericity. The ANOVA revealed a significant main effect of DCM features $\left(F(4.39, 210.64) = 228.90, MSE = 2.71, p < .001, \eta_p^2 = .83\right)$. There was no significant interaction effect between participant group and DCM features $\left(F(4.39, 210.64) = 0.83, MSE = 0.01, p = .516, \eta_p^2 = .02\right)$ and no significant main effect of group $\left(F(1, 48) = 1.12, MSE = 0.00, p = .294, \eta_p^2 = .02\right)$. This main effect of features represents the key finding as it indicates that the features in general are essential, whereas information on participant group *per se* is not.

### 3.3.    *k*-Fold Cross-Validation

To evaluate the performance of the classifier, *k*-fold CV was conducted on all the different combinations of inputs for both DCM and timeseries values.

### 3.3.1.   Cross-Validation for RCGP for Timeseries

The inputs were divided into *10* folds with 80% of the data used for training, 10% for validation, and 10% for test. After the artificial data samples were synthesised for the minor class in the training set, CV was repeated for 10 runs and the results were averaged, as shown in Table 3. Findings indicated that the Modafinil group were successfully classified from the control group with a maximum accuracy of 63.13% using RCGP in CV (minimum accuracy: 51.67%).

Table 3 - Cross-validation results for the timeseries values

| | | Training % (*SD*) | Validation % (*SD*) | Test % (*SD*) |
|---|---|---|---|---|
| Classification results for each DMN region | | | | |
| PCC | RCGP | 58.74 (5.37) | 70.36 (0.83) | 63.13 (2.86) |
| mPFC | RCGP | 57.01 (5.13) | 67.26 (3.52) | 58.69 (3.99) |
| RIPC | RCGP | 55.14 (4.91) | 68.51 (4.04) | 58.09 (5.80) |
| LIPC | RCGP | 56.97 (4.41) | 70.05 (2.41) | 58.09 (4.31) |
| Classification results for all the DMN regions (4 inputs) | | | | |
| | RCGP | 44.63 (5.99) | 71.74 (10.16) | 51.67 (8.30) |
| Classification results for all the DMN regions (1 input) | | | | |
| | RCGP | 60.21 (3.59) | 69.86 (2.53) | 60.73 (3.96) |

### 3.3.2. Cross-Validation for CGP for DCM

The inputs were divided into *10* folds with 80% of the data used for training, 10% for validation, and 10% for test. After the artificial data samples were synthesised for the minor class in the training set, CV was repeated for 10 runs and the results were averaged, as shown in Table 4. Findings revealed that the Modafinil group were successfully classified from the control group with 59.55% accuracy using CGP in CV.

Table 4 - Cross-validation results for DCM values

| | Training % (*SD*) | Validation % (*SD*) | Test % (*SD*) |
|---|---|---|---|
| CGP | 62.55 (2.87) | 76.61 (2.90) | 59.55 (6.53) |

## 4. Discussion

This research develops methods for differentiating clinical groups using EAs on rs-fMRI data, based on a controlled clinical experiment. Specifically, this experiment examined the monitoring of participants administered Modafinil (versus a control group) using rs-fMRI data. These findings build on previous research exploring the application of EAs for monitoring PD patients following treatment with Levodopa [72].

A distinctive element of this research involved the use of DCM analysis for classification. CGP classification method was used for timeseries analyses and DCM analyses and these findings were validated with two other widely used classification techniques (ANN and SVM). Findings were further validated using $k$-fold CV technique. Given that the data was highly imbalanced, ADASYN was used to balance the data before performing CV, providing a more equal class distribution within the training set (i.e., balanced numbers of Modafinil and control participants). Findings revealed a maximum classification accuracy of 75%. An important finding was that there was almost no difference in the classification accuracies between timeseries and DCM data. Moreover, EAs, specifically CGP, have not hitherto been used for classification of brain imaging data; hence, this research provided a novel application of CGP (on rs-fMRI data). CGP provided equivalent performance accuracy when compared with ANN and SVM classification methods. A relevant question for future research is whether CGP is relevant for the classification of PD patients (rather than healthy participants, as per the current research) administered Modafinil relative to non-medicated PD patients who are experiencing fatigue.

A key aspect of this research involved the application of a dynamical method of classification (RCGP). CGP has been previously used in the classification of biomedical data but not brain imaging data. This research applied CGP to rs-fMRI data, a timely approach given that fMRI is a non-invasive method that generates images with high spatial resolution and good temporal resolution and it is widely used in medical facilities [73]–[76] (e.g., for diagnosis).

fMRI data is dynamic as biomedical data from the nervous system are complex, nonlinear and nonstationary. Nevertheless, research on the classification of biomedical timeseries data typically uses static classifiers (e.g., SVM [77] and feed-forward neural networks [78]; see also [79], [80]). Other research has compared dynamical to static classifiers for biomedical data, revealing better discrimination and increased diversity in dynamical classifiers [32]. The current research used two static classifiers (SVM and ANN) to validate the findings from a dynamical classifier (RCGP) for the classification of timeseries data, which revealed comparable findings between all three methods. Whilst dynamical classifiers might theoretically better represent the nature of biomedical imaging data, future research is needed to further examine this question.

This research used a novel approach, applying classification to DCM data. Findings revealed that DCM data for classification provided comparable accuracy across all three classifiers, relative to timeseries data, which is surprising as other research has revealed that classification often provides better accuracies with raw unprocessed data. Timeseries data is raw fMRI data and includes over 100 features. In contract, DCM data is processed, representing the effective connectivity (the causal effect) of one neuronal region on another and, in this case, contained only 16 features. These findings underscore the relevance of DCM data for classification, even though this data is processed and contains much less features relative to timeseries data.

Following $k$-fold CV, classification accuracy for timeseries values decreased to 52-63% and for DCM values accuracy was reduced to 60%. This is due to the fact that this research used heavily class-imbalanced data containing 39 Modafinil participants and 111 control participants. Yet, standard classification methods typically assume balanced class distributions. Imbalanced data significantly reduces classification accuracy [56] given that the classifier cannot be trained efficiently to distinguish the differences between features in the two classes. A widely used solution involves modifying the data to obtain a sample with balanced class distributions, which often increases the overall classification accuracy compared to the original imbalanced sample [59]–[61]. Nevertheless, this solution is not perfect given that the balanced data is only used in the training set, which compromises the accuracy in the validation and test sets as the classifier that was trained for the CV is a different classifier with a different accuracy level than that used in the classification part. The current research used ADASYN to balance the training set. The fact that findings revealed accuracies of approximately up to 75% in the classification of Modafinil timeseries and DCM data speaks to the robustness of CGP as a classification method, even for highly imbalanced data as in this research.

## 5.    Conclusion

To conclude, this research explored the classification of participants administered Modafinil (versus a control group) applying both novel data (DCM values) and classification technique (CGP). This demonstrates the power of the technique to monitor PD patients in response to the medication they are receiving as their medication is adjusted. These results add to previous literature examining EAs for monitoring PD patients following Levodopa treatment [72]. Classification accuracy for DCM analyses was compared to that of timeseries analyses.

Moreover, two other classification techniques (ANN and SVM) were applied to validate the results, including employing $k$-fold CV. Findings revealed a maximum accuracy of 75% for CGP. Classification accuracy was equivalent for DCM analyses and timeseries analyses and, further, accuracy did not differ by method (CGP, ANN, or SVM). These findings underscore the relevance of CGP as a novel classification method for dynamic brain imaging data, for both processed data (as per DCM values) and raw data (i.e., timeseries). Hence, the methods developed are potentially relevant for drug treatment monitoring and for differentiating between clinical groups, for instance, the diagnosis of patients relative to healthy controls. Applying these methods developed to a clinical sample of PD patients and examining the transferability of these tools is a crucial future direction.